\documentclass[runningheads]{llncs}
\pdfoutput=1
\usepackage{amsmath}
\usepackage{amssymb}
\usepackage{color}
\usepackage{float}
\usepackage{graphicx}
\usepackage{graphicx}
\usepackage{listings}
\usepackage{subfig}
\usepackage{multirow}
\usepackage{colortbl}
\usepackage{booktabs}

\usepackage{cleveref}

\newcommand{\R}{\mathbb{R}}

\makeatletter	
\renewcommand{\paragraph}{%
  \@startsection{paragraph}{4}%
  {\z@}{0.7em}{-1em}%
  {\normalfont\normalsize\bfseries}%
}
\makeatother

\begin{document}
\pagestyle{headings}
\mainmatter

\title{Cross Pixel Optical Flow Similarity for Self-Supervised Learning} 
\titlerunning{Optical Flow Similarity Self-Supervised Learning}
\authorrunning{Mahendran, Thewlis, and Vedaldi}
\author{Aravindh Mahendran \quad James Thewlis \quad  Andrea Vedaldi}
\institute{Visual Geometry Group \\ Dept of Engineering Science, University of Oxford \\ \email{\{aravindh,jdt,vedaldi\}@robots.ox.ac.uk}}
\maketitle

\begin{abstract}
We propose a novel method for learning convolutional neural image representations without manual supervision.
We use motion cues in the form of optical flow, to supervise representations of static images.
The obvious approach of training a network to predict flow from a single image can be needlessly difficult due to intrinsic ambiguities in this prediction task.
We instead propose a much simpler learning goal: embed pixels such that the similarity between their embeddings matches that between their 
optical flow vectors.
At test time, the learned deep network can be used without access to video or flow information and transferred to tasks such as image classification, detection, and segmentation.
Our method, which significantly simplifies previous attempts at using motion for self-supervision, achieves state-of-the-art results in self-supervision using motion cues, competitive results for self-supervision in general, and is overall state of the art in self-supervised pretraining for semantic image segmentation, as demonstrated on standard benchmarks.

\end{abstract}

\section{Introduction}\label{s:intro}

Self-supervised learning has emerged as a promising approach to address one of the major shortcomings of deep learning, namely the need for large supervised training datasets.
While there is a remarkable variety of self-supervised learning methods, they are all based on the same basic premise, which is to identify problems that can be used to train deep networks without the expense of collecting data annotations.
In this spirit, an amazing diversity of supervisory signals have been proposed, from image generation to colorization, in-painting, jigsaw puzzle solving, orientation estimation, counting, artifact spotting, and many more (\cref{accv18:s:related}). Furthermore, the recent work of~\cite{doersch17} shows that combining several such cues further helps performance.

In this paper, we consider the case of \emph{self-supervision using motion cues} to learn a Convolutional Neural Network (CNN) for static images.
Here, a deep network is trained to predict, from a single video frame, how the image \emph{could change} over time.
Since predicted changes can be verified automatically by looking at the actual video stream, this approach can be used for self-supervision.
Furthermore, predicting motion may induce a deep network to learn about objects in images.
The reason is that objects are a major cause of motion regularity and hence predictability: pixels that belong to the same object are much more likely to ``move together'' than pixels that do not.

Besides giving cues about objects, motion has another appealing characteristic compared to other signals for self-supervision.
Many other methods are, in fact, based on destroying information in images (e.g.\ by removing color, scrambling parts) and then tasking a network with undoing such changes.
This has the disadvantage of learning the representation on distorted images (e.g.\ gray scale).
On the other hand, extracting a single frame from a video can be thought of as removing information only along the temporal dimension and allows one to learn the network on undistorted images.

\begin{figure}[t]
\centering
\includegraphics[width=1.0\textwidth]{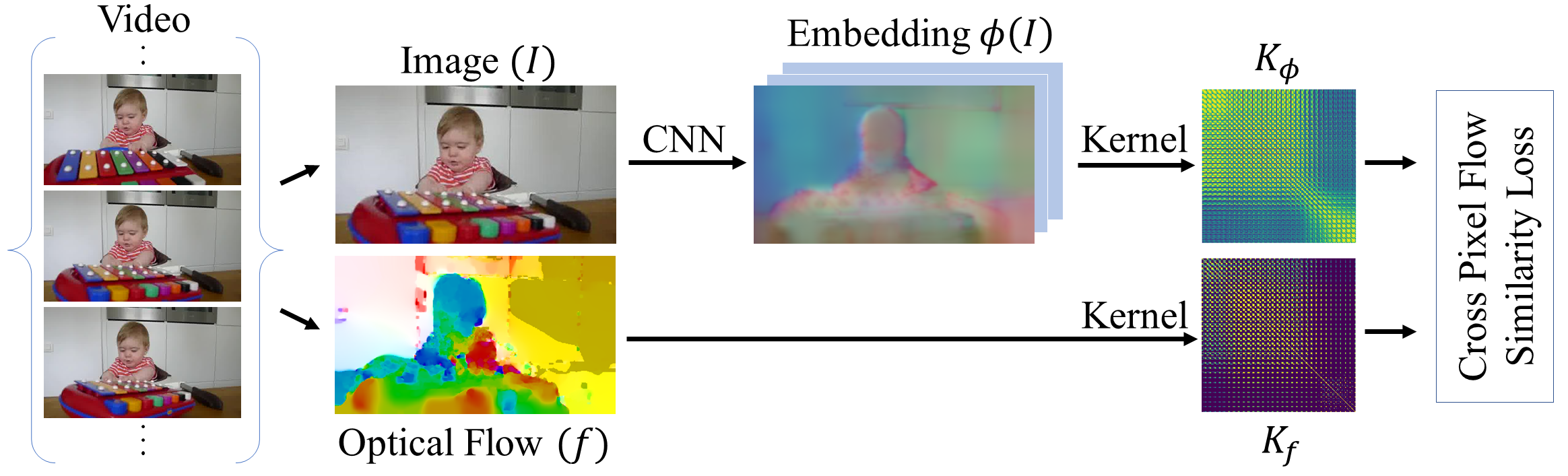} %
\caption{We propose a novel method to exploit motion information represented as optical flow, to supervise the learning of deep CNNs. We learn a network that predicts per-pixel embeddings $\phi(I)$ such that the kernel computed over these embeddings ($K_\phi$) is similar to that over corresponding optical-flow vectors ($K_f$). This allows the network to learn from motion cues while avoiding the inherent ambiguity of motion prediction from a single frame.}\label{accv18:f:method}
\end{figure}

There is however a key challenge in using motion for self-supervision: ambiguity.
Even if the deep network can correctly identify all objects in an image, this is still not enough to predict the specific direction and intensity of the objects' motion in the video, given just a single frame.
This ambiguity makes the direct prediction of the appearance of future frames particularly challenging, and overall an overkill if the goal is to learn a good general-purpose image representation for image analysis.
Instead, the previous most effective method for self-supervision using motion cues~\cite{pathak17} is based on first extracting motion tubes from videos (using off-the-shelf optical flow and motion tube segmentation algorithms) and then training the deep network to predict the resulting per-frame segments rather than motion directly.
Thus they map a complex self-supervision task into one of classic foreground-background segmentation.

While the approach of~\cite{pathak17} sidesteps the difficult problem of motion prediction ambiguity, it comes at the cost of pre-processing videos using a complex handcrafted motion segmentation pipeline, which includes many heuristics and tunable parameters.
In this paper, we instead propose a new method that can ingest cues from optical flow \emph{directly}.

Our method, presented in~\cref{s:method} and illustrated in~\cref{accv18:f:method}, is based on a new cross pixel flow similarity loss layer.
As noted above, the key challenge is that specific details about the motion, such as its direction and velocity, are usually difficult if not impossible to predict from a single frame.
We address this difficulty in two ways.
First, we learn to embed pixels into vectors that cluster together when the model believes that the corresponding pixels \emph{are likely to move together}.
This is obtained by encouraging the inner product of the learned pixel embeddings to correlate with the similarity between their corresponding optical flow vectors.
This does not require the model to explicitly estimate specific motion directions or velocities.
However, this is still not sufficient to address the ambiguity completely; in fact, while different objects may be \emph{able} to move independently, they \emph{may not do so} all the time. For example, often objects stand still, so their velocities are all zero grouping them together in optical flow.
We addressed this second challenge by using a robust loss that captures pixel grouping \emph{probabilistically} rather than deterministically.

In~\cref{accv18:s:exp} we extensively validate our model against other self-supervised learning approaches. First, we show that our approach works as well or better than~\cite{pathak17}, establishing a new state-of-the-art method for self-supervision using motion cues.
Second, to put this into context, we also compare the results to all recent approaches for self-supervision that use cues other than motion.
In this case, we show that our approach has state-of-the-art performance for semantic image segmentation.

The overall conclusion~(\cref{accv18:s:conc}) is that our method significantly simplifies leveraging motion cues for self-supervision and does better than existing alternatives for this modality; it is also competitive with self-supervision methods that use other cues, making motion a sensible choice for self-supervision by itself or in combination with other cues~\cite{doersch17}.

\section{Related Work}\label{accv18:s:related}

Self-supervised learning, of which our method is an instance, has become very popular in the community.
We discuss here the methods for training generic features for image understanding as opposed to methods with specific goals such as learning object keypoints.
We group them according to the supervision cues they use.

\paragraph{Video/Motion Based:} LSTM RNNs can be trained to predict future frames in a video~\cite{Srivastava15}.
This requires the network to understand image dynamics and extrapolate it into the future. However, since
several frames are observed simultaneously, these methods may learn something
akin to a tracker, with limited abstraction. 
On the other hand, we learn to predict properties of 
optical flow from a \textbf{single input image}, thus learning a static image representation rather 
than a dynamic one. Closely related to our work is the use of \emph{video segmentation} 
by~\cite{pathak17}. They use an off-the-shelf video segmentation method~\cite{faktor14}
to construct a foreground-background segmentation dataset in an unsupervised manner.
A CNN trained on this proxy task transfers well when fine-tuned for object recognition
and detection. We differ from them in that we do not require a sophisticated pre-existing
pipeline to extract video segments, but use optical flow directly. Also closely related to us
is the work of~\cite{agrawal15}. They train a Siamese style convolutional neural network
to predict the transformation between two images. The individual base networks in their
Siamese architecture share weights and can be used as feature extractors for single
images at test time. This late fusion strategy forces the learning of abstractions, but
our \textbf{no-fusion approach} pushes the model even further to learn better features.
The polar opposite of these is to do early fusion by concatenating two frames as in
FlowNet~\cite{Dosovitskiy15}. This was used as a pretraining strategy by~\cite{gan18}
to learn representations for \textbf{pairs of frames}. This is different from our
objective as we aim to learn a \textbf{static image representation}. This difference
becomes clearer when looking at the evaluation. While we evaluate on image
classification, detection, and segmentation;~\cite{gan18} evaluate on dynamic scene
and action recognition.

Temporal context is a powerful signal. \cite{misra16,wei18,Lee17}~learn to predict the
correct ordering of frames. \cite{isola15}~exploit both temporal and spatial co-occurrence
statistics to learn visual groups. \cite{jayaraman16}~extend slow feature analysis using 
higher order temporal coherence. \cite{wang15a}~track patches in a video to supervise their
embedding via a triplet loss while~\cite{gao16} do the same but for spatio-temporally matched
region proposals. Temporal context is applied in the imitation learning setting by Time 
Contrastive Networks~\cite{sermanet18}.

Videos contain more than just temporal information. Some methods exploit audio channels by
prediction audio from video frames~\cite{de94,owens16}. \cite{Arandjelovic17}~train a two
stream architecture to classify whether an image and sound clip go together or not. 
Temporal information is coupled with ego-motion in~\cite{jayaraman17,jayaraman15}.
\cite{wang17}~use videos along with spatial context pretraining~\cite{doersch15a} to 
construct an image graph. Transitivity in the graph is exploited to learn representations
with suitable invariances.

\paragraph{Colorization:} \cite{larsson16,larsson17,zhang16}~predict colour information given 
greyscale input and show competitive pre-training performance.
A generalization to arbitrary pairs of modalities was proposed in~\cite{zhang17}. 
 
\paragraph{Spatial Context:} \cite{pathak16}~solve the in-painting problem, where a network is tasked 
with filling-in partially deleted parts of an image.
\cite{doersch15a}~predict the relative position of two patches extracted from an image.
In a similar spirit, \cite{noroozi16,noroozi18}~solve a jigsaw puzzle problem.
\cite{noroozi18}~also cluster features from a pre-trained network to generate pseudo labels, 
which allows for knowledge distillation from larger networks into smaller ones.
The latest iteration on context prediction by~\cite{mundhenk17} obtains state-of-the-art results on several benchmarks.
 
\paragraph{Adversarial/Generative:} BiGAN based pretrained models~\cite{Donahue17} show competitive performance on various recognition benchmarks.
\cite{jenni18}~adversarially learn to generate and spot defects.
\cite{ren18}~obtain self-supervision from synthetic data and adapt their model to the domain of real images by training against a discriminator.
\cite{bojanowski17}~predict noise-as-targets via an assignment matrix which is optimized on-line.
Their approach is domain agnostic.
More in general, generative unsupervised layer-wise pretraining was extensively used in deep learning before AlexNet~\cite{Krizhevsky12}.
An extensive review of these and more recent unsupervised generative models is beyond the scope of our paper.
 
\paragraph{Transformations:} \cite{Dosovitskiy16a}~create surrogate classes by applying a set of transformations to each image and learn to become invariant to them.
\cite{Gidaris18}~do the opposite and try to estimate the transformation (just one of four rotations in their case) given the transformed image.
The crop-concatenate transformation  is implicit in the learning by counting method of~\cite{noroozi17}.
\cite{novotny18}~use correspondences obtained from synthetic warps to learn a dense image representation.
 
\paragraph{Others:} A combination of self-supervision approaches was explored by~\cite{doersch17}.
They report results only with ResNet models making it hard to compare with concurrent work, but 
closely matching ImageNet-pretrained networks in performance on the PASCAL VOC detection task.
\cite{zhan18}~propose a mix-and-match tuning strategy as a precursor to finetuning on the target domain.
Their approach can be applied to any pretrained model and achieves impressive results for PASCAL VOC 2012 semantic segmentation.
Another widely-applicable trick that helps in transfer learning is the re-balancing method of~\cite{krahenbuhl16}.
Lastly, our optical-flow classification baseline is based on the work of~\cite{bansal17}.
They learn a sparse hypercolumn model to predict surface normals from a single image and use this as a pretraining strategy.
Our baseline flow classification model is the same but with AlexNet for discretized optical-flow.  
\section{Method}\label{s:method}

In this section, we describe our novel method, illustrated in~\cref{accv18:f:method}, for self-training deep neural networks via direct ingestion of optical flow.
Once learned, the resulting image representation can be used for classification, detection, segmentation and other tasks with minimal supervision.

Our goal is to learn the parameters $\Theta$ of a neural network $\phi$ that maps a single image or frame $I : \R^2 \supset \Omega \rightarrow \R^3$ to a field of pixel embeddings $\phi(I, p | \Theta) \in \R^D$, one for each pixel $p \in \Omega$.
In order to learn this embedding, which is extracted from a \emph{single frame}, we task our neural 
network with \emph{predicting} the motion present in the corresponding video, represented as optical flow.
However, since predicting flow vectors directly is ambiguous, we propose instead to require the \emph{similarity} between \emph{pairs} of embedding vectors to align to the similarity between the corresponding flow vectors.
This is sufficient to capture the idea that things that move together should be grouped together, 
popularly known as the Gestalt's principle of \emph{common fate}.

Formally, given $D$-dimensional CNN embedding vectors $\phi(I, p | \Theta), \phi(I, q | \Theta) \in \R^D$ for pixels $p,q \in \Omega$ and their corresponding flow vectors $f_p,f_q\in\mathbb{R}^2$, we match the kernel matrices
\begin{equation}
\forall p,q \in \Omega: \quad
K_\phi\Big (\phi(I, p | \Theta), \phi(I, q | \Theta) \Big ) \approxeq K_f(f_p, f_q)
\label{accv18:eqn:kernelalign}
\end{equation}
where $K_\phi: \R^D \times \R^D \rightarrow \R$, $K_f: \R^2 \times \R^2 \rightarrow \R$ are kernels that measure the similarity of the CNN embeddings and flow vectors, respectively.

In this formulation, in addition to the choice of CNN architecture $\phi$, the key design decisions are the choice of kernels $K_\phi, K_f$ and how to translate constraint~\eqref{accv18:eqn:kernelalign} into a loss function.
The rest of the section discusses these choices.

\paragraph{Kernels:}
\begin{figure}[t]
\centering
\includegraphics[width=1.0\textwidth]{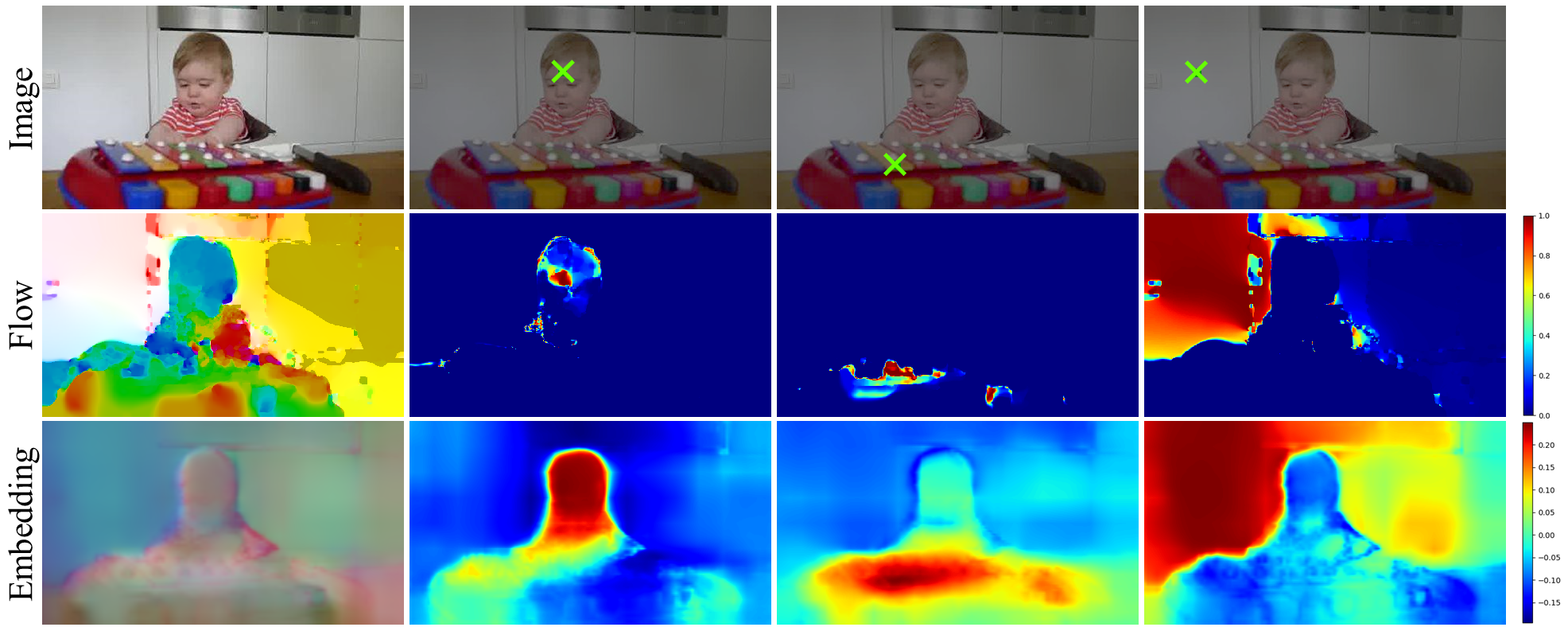}

\caption{Visualization of flow (second row) and embedding (third row) kernels. For three pixels $p$, we plot the row $K(p, \cdot)$ reshaped as an image, showing which pixels go together from the kernel's perspective. Note the localized nature of the flow kernel which is obtain by setting a low bandwidth for the RBF kernel. In the first column, optical flow and embeddings (after a random $16D \rightarrow 3D$ projection) are visualized as color images.}\label{accv18:f:kernelvis}%
\end{figure}
In order to compare CNN embedding vectors and flow vectors, we choose respectively the (scaled) cosine similarity kernel and the Gaussian/RBF kernel.
Using the shorthand notation $\phi_p = \phi(I, p | \Theta)$ for readability, these are:
\begin{equation}
K_\phi(\phi_p, \phi_q) := \frac{1}{4} \frac{\phi_p^T \phi_q}{\|\phi_p\|_2 \|\phi_q\|_2},
\qquad
K_f(f_q, f_q) := \exp\left(
    -\frac{\|f_p - f_q\|_2^2}{2\sigma^2}
\right).
\label{accv18:e:kernels}
\end{equation}
Note that these kernels, when restricted to the set of pixels $\Omega$, are matrices of size $|\Omega| \times |\Omega|$.
Each row or column of this matrix can be thought of as a heatmap capturing the similarity of a given pixel with respect to all other pixels and thus can be visualized as an image.
We present such visualizations for both of our kernels in~\cref{accv18:f:kernelvis}.

We use the Gaussian kernel for the flow vectors as this is consistent with the Euclidean interpretation of optical flow as a displacement.
Observe that reducing kernel bandwidth $(\sigma)$ results in a localized kernel that pushes our embeddings harder to distinguish between different movable objects. The value of $\sigma$ is learned along with the weights of the CNN in the optimization. 
This localized kernel, with learned $\sigma^2=0.0036$, is shown in the second row of~\cref{accv18:f:kernelvis}.

We use the cosine kernel for the learned embedding as the CNN effectively computes a \emph{kernel feature map}, so that in principle it can approximate any kernel via the inner product.
However, note that the expression normalizes vectors in $L^2$ norm, so that this inner product is the cosine of the angle between embedding vectors. This normalization is key as it guarantees that $K_\Phi$ is maximum when the embeddings being compared are identical (the Gaussian kernel does so automatically).

\paragraph{Cross Pixel Optical-Flow Similarity Loss function:}

The constraint in equation~\eqref{accv18:eqn:kernelalign} requires kernels $K_\phi$ and $K_f$ to be similar.
A conventional metric to measure the similarity between kernels is \emph{kernel target alignment} (KTA)~\cite{Cristianini00} which, for two kernel matrices $K,K'$, is given by $\sum_{pq} K_{pq} K'_{pq} / \sqrt{\sum_{pq} K_{pq}^2 \sum_{pq} {K'_{pq}}^2}$.
In this manner, KTA and analogous metrics require kernels to match \emph{everywhere}.
In our case, however, this is too strong a requirement.
Even if the embedding function $\phi$ correctly groups pixels that belong to different movable objects, $K_\phi$ may still not match the optical flow kernel $K_f$.
The reason is that in many video frames objects may not move with a distinctive pattern or may not move at all, so that no corresponding grouping can be detected in the measured flow field.
KTA failed to produce reasonable embeddings in our preliminary experiments.

This motivated us to research a kernel similarity criterion more suitable for our task.
The key idea is to relax the correspondence between kernels to hold \emph{probabilistically}.
We do so in two steps.
First, we re-normalize each column $K_*(\cdot, q)$ of each kernel matrix into a probability distribution $S_*(\cdot, q)$.
Distribution $S_f(\cdot|q)$ describes which image pixels $p$ are likely to belong to the same segment as pixel $q$, according to optical flow. $S_\phi(\cdot|q)$ describes the same but from the CNN embedding's perspective.
These distributions, arising from CNN and optical flow kernels, are compared by using cross entropy, summed over columns:
\begin{equation}
\mathcal{L}(\Theta) = -\sum_q \sum_p S_f(p, q) \log S_\phi(p, q).
\end{equation}

Normalization uses the softmax operator. We reduce the contribution of diagonal terms in the kernel matrix before this normalization because each pixel is trivially similar to itself and would skew the softmax normalization.
Formally, for optical flow we have:
\begin{equation}
S_f(p, q)
=
\begin{cases}
    \frac{1}{\sum_{q' \ne p} \left. \exp(K_f(p, q')) \right. + 1}, & \mbox{if } p = q, \\
    \frac{\exp(K_f(p, q'))}{\sum_{q' \ne p} \left. \exp(K_f(p, q')) \right. + 1}, & \mbox{if } p \ne q.
\end{cases}
\end{equation}
Similarly, for the CNN embedding we have:
\begin{equation}
S_\phi(p, q)
=
\begin{cases}
\frac{e^{-3/4}}{\sum_{q' \ne p} \left. \exp(K_\phi(p, q')) \right. + e^{-3/4}}, & \mbox{if } p = q, \\
\frac{\exp(K_\phi(p, q'))}{\sum_{q' \ne p} \left. \exp(K_\phi(p, q')) \right. + e^{-3/4}}, & \mbox{if } p \ne q.
\end{cases}
\end{equation}
Note that the $p = q$ and $p = q'$ cases contribute $e^{-3/4}$ for the embedding kernel's softmax. This is because of the $1/4$ scaling used in the cosine similarity kernel~\eqref{accv18:e:kernels}.

\paragraph{CNN embedding function:}

Lastly, we discuss the architecture of the CNN function $\phi$ itself.
We design the embedding CNN as a hypercolumn head~\cite{hariharan15} over a conventional CNN backbone such as AlexNet.
The hypercolumn concatenates features from multiple depths so that our embedding can exploit high resolution details normally lost due to max-pooling layers.
For training, we use the sparsification trick of~\cite{larsson16} and restrict prediction and loss computation to a few randomly sampled pixels in every iteration.
This reduces memory consumption and improves training convergence as pixels in the same image are highly correlated and redundant; via sampling we can reduce this correlation and train more efficiently~\cite{bansal17}.

In more detail, the backbone is a CNN with activations at several layers:
$\{\phi_{c_1}(I | \Theta), \cdots,$ $\phi_{c_n}(I | \Theta)\} \in \R^{H_1 \times W_1 \times D_1} \times \cdots \times \R^{H_n \times W_n \times D_n}$.
We follow~\cite{larsson17} and interpolate values at a given pixel location and concatenate them to form a hypercolumn $\phi_H(I, p | \Theta) \in \R^{D_1 + \cdots + D_n}$.
The hypercolumn is then projected non-linearly to the desired embedding $\phi(I, p | \Theta) \in \R^D$ using a multi-layer perceptron (MLP). 
Details of the model architecture are discussed in~\cref{accv18:s:backbone}.

\section{Experiments}\label{accv18:s:exp}

We extensively assess our approach by demonstrating its effectiveness in learning features that we show as useful for several tasks.
In order to make our results comparable to most of the related papers in the literature, we consider an AlexNet~\cite{Krizhevsky12} backbone and four tasks: classification in ImageNet~\cite{Russakovsky15short} and classification, detection, and segmentation in PASCAL {VOC}~\cite{pascal07,pascal-voc-2012}.

\subsection{Backbone details}\label{accv18:s:backbone}

We adapt the AlexNet version used by Pathak et al.~\cite{pathak17}. The modifications are minor (mostly related to padding) and to make it suitable to attach a hypercolumn head.
Sparse hypercolumns are built from the conv1, pool1, conv3, pool5 and fc7 AlexNet activations.
Embeddings are generated using a multi-layer perceptron (MLP) with a single hidden layer and are L2-normalized. The embeddings are $D=16$ dimensional (this number could be improved via cross validation, although this is expensive). The exact model specification, as a caffe .prototxt, is included in supplementary material.

\subsection{Dataset}\label{accv18:s:dataset}
\begin{figure}[t]
   \centering
   \includegraphics[width=0.95\textwidth]{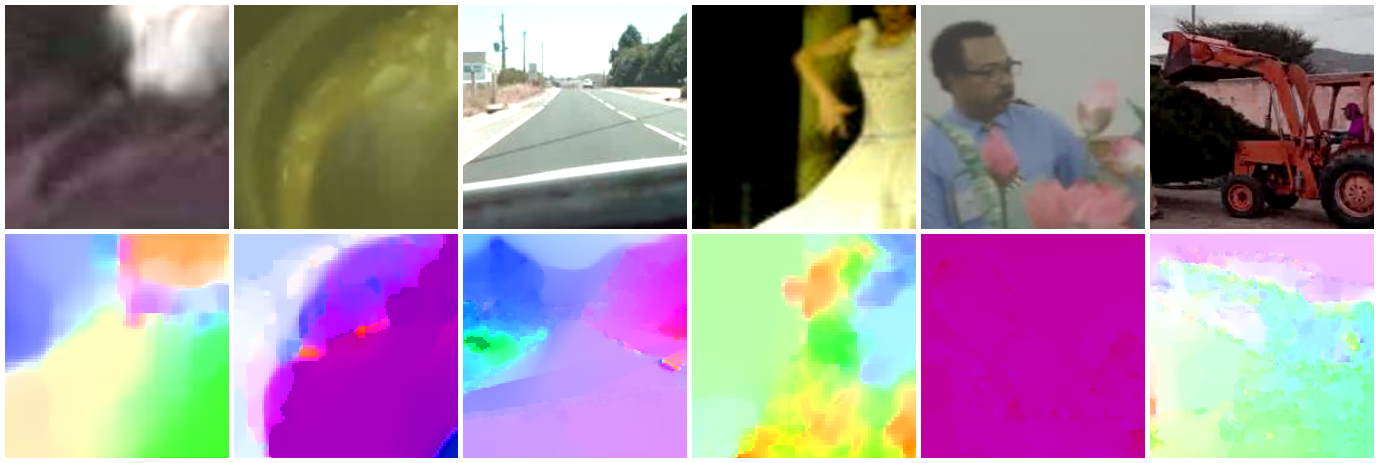}

   \vspace{-0.1em}
   \caption{Image and optical-flow training pairs post scale-crop-flip data augmentation. The noisy nature of both images and optical-flows illustrate the challenges in using motion as a self-supervision signal.}%
   \label{accv18:f:traindata}%
\end{figure}
We train the above AlexNet model on a dataset of RGB-optical flow image pairs. 
Inspired by the work of Pathak et al.~\cite{pathak17}, we built a dataset from  $\sim 204k$ videos in the YFCC100m dataset~\cite{thomee16}.
The latter consists of Flickr videos made publicly-available under the creative commons license.
We extract 8 random frames from each video and compute optical flow between those at times $t$ and $t+5$ using the same (handcrafted) optical flow method of~\cite{pathak17,Liu09b}.
Overall, we obtain 1.6M image-flow pairs.\footnote{%
The dataset occupies more than 1TB of space and does not easily fit in fast memory such as an SSD for training.
We addressed this problem by using a fixed point 16bit representation of optical flow similar to KITTI~\cite{Geiger12} and storing it as PNG images, with dramatic compression and negligible residual error. The compressed data occupies 431GB.}
Example training sample crops along with optical-flow fields are shown in~\cref{accv18:f:traindata}.
The noisy nature of both the images and optical-flow in such large-scale non-curated video collections makes it all the more challenging for self-supervision.

Optical flow vectors $(f_x, f_y)$ are normalized logarithmically to lie between $[-1,1]$ during training, so that occasional large flows do not dominate learning. 
More precisely, the normalization is given by:
\begin{equation}\label{accv18:eqn:flownormalization}
f^\prime = \left[ \begin{array}{c} 
\operatorname{sign}(f_x) \min \left(1,\frac{\log(|f_x|+1)}{\log(M+1)} \right) \\
\vspace{0.01em} \\
\operatorname{sign}(f_y) \min \left(1,\frac{\log(|f_y|+1)}{\log(M+1)} \right) \end{array} \right]
\end{equation}
where $M$ is a loose upper bound on the flow-magnitude set to $56.0$ in our experiments.

Despite the large size of this data and aggressive data augmentation during training, AlexNet overfits on our self-supervision task.
We use early stopping to reduce over-fitting by monitoring the loss on a validation set.
The validation set consists of 5000 image-flow pairs computed from the YouTube objects dataset~\cite{prest2012learning}.
Epic-Flow~\cite{revaud2015epicflow} + Deep-Matching~\cite{Weinzaepfel13} was used to compute optical-flow for these frames.

\subsection{Learning Details}

AlexNet is trained using the Adam optimizer~\cite{kingma14} with $\beta_1=0.9, \beta_2=0.999, \epsilon=10^{-8}$ and initial learning rate set to $10^{-4}$.
No weight decay is used because it worsened the minima reached by our models before overfitting started.
Pixels are sampled uniformly at random for the sparse hypercolumns.
Sampling more pixels gives more information per image but also consumes more memory and is computationally expensive making each iteration slower.
We use 512 pixels per image to balance this trade-off.
This reduces memory consumption and allows for a large batch size of 96 frames.
Scale, horizontal flip and crop augmentation are applied during training. 
The network is trained on crops of size $224 \times 224$. 
Parameter-free batch-normalization~\cite{Ioffe15} is used throughout the network; the moving average mean and variance are absorbed into convolution kernels after self-supervised training, so that, for evaluation, AlexNet does not contain batch normalization layers.
The full implementation using TensorFlow~\cite{Abadi16} will be released to the public upon acceptance.

\subsection{Qualitative Results}
While our learned pixel embeddings are not meant to be used directly (instead their purpose is to pre-train a neural network parametrization that can be transferred to other tasks), nevertheless it is informative to visualize them.

\begin{figure}[t!]
\centering
\includegraphics[width=1.0\textwidth]{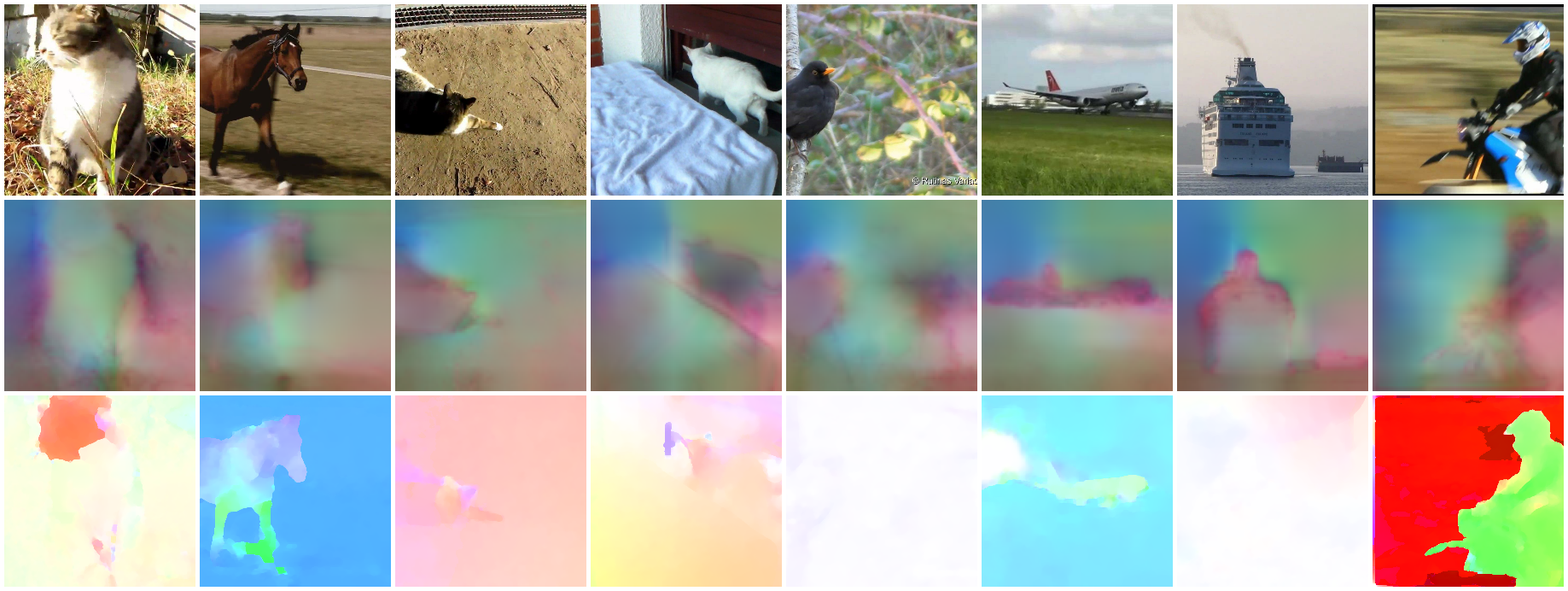}

\caption{Per-pixel embeddings are visualized by randomly projecting them to RGB colors. We show example embeddings generated by our model on frames from the YouTube objects dataset. From top to bottom: The original validation images, RGB-mapped embeddings, and optical-flow fields. Best viewed in color on screen.}\label{accv18:f:embeddingvis}%
\end{figure}

\paragraph{Embedding Visualizations:} 
Since embeddings are 16D, we first project them to 3D vectors via random projections and them map the resulting coordinates to RGB space.
We show results on the YouTube objects validation set in~\cref{accv18:f:embeddingvis}.
Note that pixels on a salient foreground object tend to cluster together in the embedding (see, for example, the cats in columns 1, 3 and 4, the aircraft in column 6 and the motor-cyclist in column 8).

\begin{figure}[t!]
   \centering
   \includegraphics[width=5cm,clip,trim=0 0 0 12.5cm]{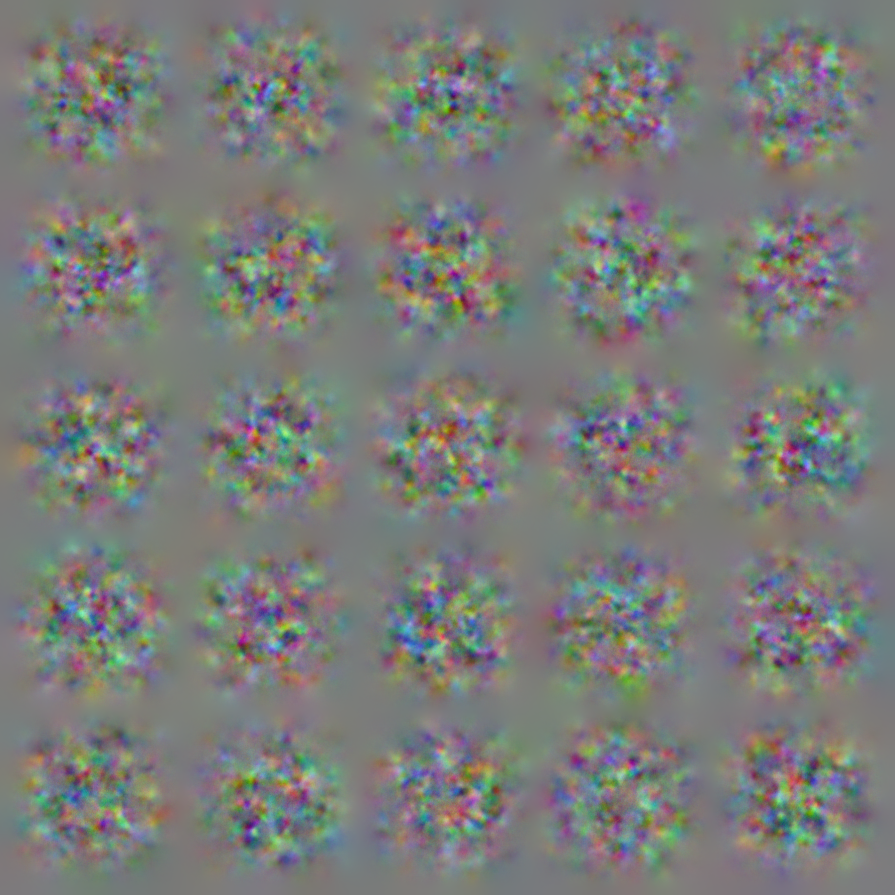} ~~~~~ %
   \includegraphics[width=5cm,clip,trim=0 0 0 12.5cm]{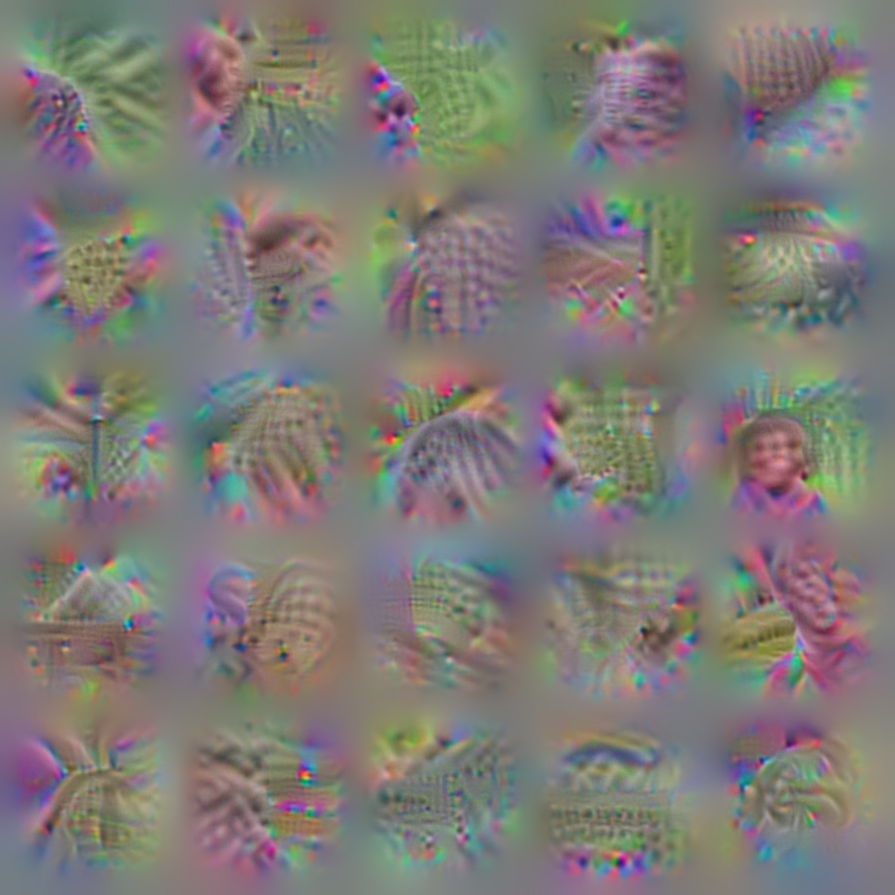}

\vspace{-0.5em}
\caption{Each image in the grid corresponds to the estimated optimal excitatory stimulus for a neuron in conv5 feature space. This visualization is obtained using the regularized neuron maximization method of~\cite{Mahendran16}. Left: Neurons in a randomly initialized AlexNet. Right: Neurons in AlexNet trained using our approach: significantly more structure emerges.}\label{accv18:f:neuronmax}%
\end{figure}

\paragraph{Neuron Maximization:}
We use per-neuron activation maximization~\cite{Mahendran16} to visualize individual neurons in the fifth convolutional layer (\cref{accv18:f:neuronmax}).
This presents the estimated optimal stimulus for each of these neurons, made interpretable using a natural image prior.
We observe abstract patterns including a human form (row 1, column 5) that are obviously not present in a random network,
suggesting that the representation may be learning concepts useful for general-purpose image analysis.

\subsection{Quantitative results}
\setlength{\tabcolsep}{4pt}
\begin{table}[ht]
  \centering
  \caption{Pascal VOC Comparison for three benchmarks: VOC2007-classification (column 4) \%mAP, VOC2007-Detection (Column 5) \%mAP and VOC2012-Segmentation (Column 6) 
  {\%mIU}. The rows are grouped into four blocks (0) The limits of no-supervision and human supervision, (1) motion/video based self-supervision, (2) Our model and the baseline, (3) others. The third column [ref] indicates which publication the reported numbers are borrowed from. Zhan et al. (marked $^\dagger$) contribute a different mix-and-match tuning strategy which transfers better to the target domain compared to finetuning. This is orthogonal to the efforts of finding a good self-supervision method and is therefore not counted when marking the state-of-the-art in \textbf{bold.}}\label{accv18:t:pascalvoc}
  \footnotesize
  \begin{tabular}{c | l l l | c c c}
  \toprule
  &Method & Supervision & [Ref] & Cls. & Detection & Seg. \\
  \midrule
  &Krizhevsky et al.~\cite{Krizhevsky12} & Class Labels & \cite{zhang16} & 79.9 & 56.8 & 48.0 \\
  &Random & - & \cite{pathak16} & 53.3 & 43.4 & 19.8 \\
  \midrule
  \cellcolor[gray]{.9}&Agrawal et al.~\cite{agrawal15} & Egomotion & \cite{Donahue17} & 63.1 & 43.9   & - \\
  \cellcolor[gray]{.9}&Jayaraman et al.~\cite{jayaraman15}  & Egomotion & \cite{jayaraman15} & - & 41.7 & -  \\
  \cellcolor[gray]{.9}&Lee et al.~\cite{Lee17} & Time-order & \cite{Lee17} & 63.8 & 46.9 & - \\
  \cellcolor[gray]{.9}&Misra et al.~\cite{misra16} & Time-order & \cite{misra16} & - & 42.4 & - \\
  \cellcolor[gray]{.9}&Pathak et al.~\cite{pathak17} & Video-seg & \cite{pathak17},Self & 61.0 & 50.2 & - \\
  \cellcolor[gray]{.9}&Wang et al.~\cite{wang15a} & Track + Rank & \cite{krahenbuhl16,wang15a} & 63.1 & 47.5 & - \\
  \cline{2-7}
  \cellcolor[gray]{.9}&Ours & Optical-flow & Self & \textbf{64.4} & \textbf{50.3} & \textbf{41.4}\\
  \cellcolor[gray]{.9}\multirow{-8}{*}{\rotatebox[origin=c]{90}{Motion cues}}
  &Ours direct reg.\ & Optical-flow & Self & 61.4 & 44.0 & 37.1\\
  \midrule
  \midrule
  \parbox[t]{2mm}{\multirow{18}{*}{\rotatebox[origin=c]{90}{Other cues}}}
  &Bojanowski et al.~\cite{bojanowski17} & - & \cite{bojanowski17} & 65.3 & 49.4 & - \\
  &Doersch et al.~\cite{Doersch15} & Context & \cite{Donahue17} & 65.3 & 51.1 & - \\
  &Donahue et al.~\cite{Donahue17} & - & \cite{Donahue17} & 60.3 & 46.9 & 35.2 \\
  &Gidaris et al.~\cite{Gidaris18} & Rotation & \cite{Gidaris18} & \textbf{73.0} & 54.4 & 39.1 \\
  &Krahenbuhl et al.~\cite{krahenbuhl16} & - & \cite{krahenbuhl16,Donahue17} & 56.6 &  45.6 & 32.6\\
  &Larssons et.al.~\cite{larsson17} & Colorization & \cite{larsson17} & 65.9 & - & 38.4\\
  &Mundhenk et al.~\cite{mundhenk17} & Context & \cite{mundhenk17} & 69.3 & 55.2 & \textbf{40.6} \\
  &Noroozi et al.~\cite{noroozi16} & Jigsaw & \cite{noroozi16} & 67.6 &  53.2 & 37.6 \\
  &Noroozi et al.~\cite{noroozi17} & Counting & \cite{noroozi17} & 67.7 & 51.4 & 36.6 \\
  &Noroozi et al.~\cite{noroozi18} & Jigsaw++ & \cite{noroozi18} & 69.8 & \textbf{55.5} & 38.1 \\
  &Noroozi et al.~\cite{noroozi18} & CC+Jigsaw++ & \cite{noroozi18} & 69.9 & 55.0 & 40.0 \\
  &Owens et al.~\cite{owens16} & Sound & \cite{zhang17,owens16} &  61.3 & 44.1 & - \\
  &Pathak et al.~\cite{pathak16} & In-painting & \cite{pathak16} & 56.5 & 44.5 & 29.7 \\
  &Jenni et al.~\cite{jenni18} & - & \cite{jenni18} & 69.8 & 52.5 & 38.1 \\
  &Zhang et al.~\cite{zhang16} & Colorization & \cite{zhang16} & 65.9 & 46.9 & 35.6 \\
  &Zhang et al.~\cite{zhang17} & Split-Brain & \cite{zhang17} & 67.1 & 46.7 & 36.0 \\
  \cmidrule{2-7}
  &Zhan et al.~\cite{zhan18}$^\dagger$ & Colorization & \cite{zhan18} & - & - & 42.8 \\
  &Zhan et al.~\cite{zhan18}$^\dagger$ & Puzzle & \cite{zhan18} & - & - & 44.5 \\
  \bottomrule
  \end{tabular}  
\end{table}
\setlength{\tabcolsep}{1.4pt}
\setlength{\tabcolsep}{3pt}
\begin{table}[ht]
  \centering
  \caption{ImageNet LSVRC-12 linear probing evaluation. A linear classifier is trained on the (downsampled) activations of each layer in the pretrained model. Top-1 accuracy is reported on ILSVRC-12 validation set. The column ``[ref]'' indicates which publication the reported numbers are borrowed from. We finetune Pathak et.al.'s model along with ours as they do not report these benchmark in their paper.}\label{accv18:t:ilsvrclinear}
  \footnotesize
  \begin{tabular}{c | l l l | c c c c c }
  \toprule
  &Method & Supervision & [ref] & Conv1 & Conv2 & Conv3 & Conv4 & Conv5 \\
  \midrule
  &Krizhevsky et.al.~\cite{Krizhevsky12} & Class Labels & \cite{zhang17} & 19.3 & 36.3 & 44.2 & 48.3 & 50.5 \\
  &Random & - & \cite{zhang17} & 11.6 & 17.1 & 16.9 & 16.3 & 14.1 \\
  &Random-rescaled~\cite{krahenbuhl16} & - & \cite{krahenbuhl16} & 17.5 & 23.0 & 24.5 & 23.2 & 20.6 \\
  \midrule  
  \cellcolor[gray]{.9}&Pathak et.al.~\cite{pathak17} & Video-seg & Self & \textbf{15.8} & 23.2 & \textbf{29.0} & \textbf{29.5} & 25.4 \\
  \cellcolor[gray]{.9}&Ours & Optical-Flow & Self & 15.0 & \textbf{24.8} & 28.9 & 29.4 & \textbf{28.0} \\
  \multirow{-3}{*}{\rotatebox[origin=c]{90}{Motion}}
  \cellcolor[gray]{.9}&Ours direct reg.\ & Optical-Flow & Self & 14.0 & 22.6 & 25.3 & 25.0 & 23.0 \\
  \midrule
  \midrule
  \multirow{11}{*}{\rotatebox[origin=c]{90}{Other cues}}
  &Doersch et.al.~\cite{doersch15a} & Context & \cite{zhang17}     & 16.2          & 23.3          & 30.2          & 31.7          & 29.6 \\
  &Gidaris et.al.~\cite{Gidaris18} & Rotation & \cite{Gidaris18}   & 18.8          & 31.7          & \textbf{38.7} & 38.2          & \textbf{36.5} \\
  &Jenni et.al.~\cite{jenni18} & - & \cite{jenni18}                & 19.5          & \textbf{33.3} & 37.9          & \textbf{38.9} & 34.9 \\
  &Mundhenk et.al.~\cite{mundhenk17} & Context & \cite{mundhenk17} & \textbf{19.6} & 31.4          & 37.0          & 37.8          & 33.3 \\
  &Noroozi et.al.~\cite{noroozi16} & Jigsaw & \cite{noroozi17}     & 18.2          & 28.8          & 34.0          & 33.9          & 27.1 \\
  &Noroozi et.al.~\cite{noroozi17} & Counting & \cite{noroozi17}   & 18.0          & 30.6          & 34.3          & 32.5          & 25.7 \\
  &Noroozi et.al.~\cite{noroozi18} & Jigsaw++ & \cite{noroozi18}   & 18.2          & 28.7          & 34.1          & 33.2          & 28.0 \\
  &Noroozi et.al.~\cite{noroozi18} & CC+Jigsaw++ & \cite{noroozi18}& 18.9          & 30.5          & 35.7          & 35.4          & 32.2 \\
  &Pathak et.al.~\cite{pathak16} & In-Painting & \cite{zhang17}    & 14.1          & 20.7          & 21.0          & 19.8          & 15.5 \\
  &Zhang et.al.~\cite{zhang16} & Colorization & \cite{zhang17}     & 13.1          & 24.8          & 31.0          & 32.6          & 31.8 \\
  &Zhang et.al.~\cite{zhang17} & Split-Brain & \cite{zhang17}      & 17.7          & 29.3          & 35.4          & 35.2          & 32.8 \\
  \bottomrule
  \end{tabular}  
\end{table}
\setlength{\tabcolsep}{1.4pt}
We follow the standard practice in the self-supervised learning community and fine-tune the learned representation on various recognition benchmarks.
We evaluate our features on the PASCAL VOC 2007 detection and classification~\cite{pascal07}, PASCAL VOC 2012 segmentation~\cite{pascal-voc-2012}, and ILSVRC12 linear probing benchmarks~\cite{zhang16} (in the latter case, the representation is frozen).
We provide details on the evaluation protocol next and compare against other self-supervised models with results reported for AlexNet like architectures
in~\cref{accv18:t:pascalvoc} and~\cref{accv18:t:ilsvrclinear}.
Differently from other approaches, we did not benefit from the re-balancing trick of~\cite{krahenbuhl16} and report results without it.

\paragraph{Baseline:}

Our main hypothesis is that our similarity matching method, rather than a direct prediction of optical flow, is a more appropriate mechanism for exploiting optical flow information as a supervisory signal.
We validate this hypothesis by comparing against a direct optical-flow prediction baseline, using the same CNN architecture but a different loss function: while we use a flow-similarity matching loss, this baseline does a standard per-pixel softmax cross entropy across 16 discrete optical flow classes, once for each spatial dimension --- $x$ and $y$. 
To this end, since the flow is normalized in $[-1, 1]$ (\cref{accv18:eqn:flownormalization}), this interval is discretized uniformly.
Note that direct $L^2$ regression of flow vectors is also possible, but did not work as well in preliminary experiments. This may be because continuous regression is usually harder for deep networks compared to classification especially for ambiguous tasks.
It was beneficial to use a faster initial learning rate of $0.01$.

\paragraph{VOC2007-Detection:}

We finetune our AlexNet backbone end-to-end using the Fast-RCNN model~\cite{Girshick15} to obtain results for PASCAL VOC 2007 detection~\cite{pascal07}.
Finetuning is performed for 150k iterations, were the learning rate drops by a $10^{th}$ every 50k iterations.
The initial learning rate is set to $0.002$, weight decay to $5 \times 10^{-4}$, train-set is VOC2007-train+val, test-set is VOC2007-test.
Following the guidelines of~\cite{krahenbuhl16}, we use multi-scale training  and single scale testing.
We report mean average precision (mAP) in~\cref{accv18:t:pascalvoc} (col.~5) along with results of  other self-supervised learning methods.
We are comparable to the state-of-the-art among methods that use temporal information in videos for self-supervision.

\paragraph{VOC2007 classification:}

We finetune our pretrained AlexNet to minimize the softmax cross-entropy loss over the PASCAL VOC 2007 \textit{trainval} set.
The initial learning rate is $10^{-3}$ and drops by a factor of 2 every 10k iterations for a 
total of 80k iterations and predictions are averaged over 10 random crops at test time in keeping 
with~\cite{krahenbuhl16}.
We use the code provided by~\cite{larsson17} and report mean average precision on VOC2007-test in the fourth column of~\cref{accv18:t:pascalvoc}.
We achieve state-of-the-art among methods that derive self-supervision from motion cues; in particular, we outperform~\cite{pathak17} by a large margin.

\paragraph{ILSVRC12 linear probing:}

We follow the protocol and code of~\cite{zhang17} to train a linear classifier on activations of our pre-trained network.
The activation tensors produced by various convolutional layers (after ReLU) are down-sampled using bilinear interpolation to have roughly 9,000-10,000 elements before being fed into a linear classifier.
The CNN parameters are frozen and only the linear classifier weights are trained on the ILSVRC-12 training set.
Top-1 classification accuracy is reported on the ILSVRC-12 validation set (\cref{accv18:t:ilsvrclinear}).
We perform comparably to the best motion-based self-supervision method of~\cite{pathak17} (slightly worse or better depending on the layer), but using other types of cues achieves stronger results in this case.

\paragraph{VOC2012 segmentation:}

We use the two staged finetuning approach of~\cite{larsson17} who finetune their AlexNet for semantic segmentation using a sparse hypercolumn head instead of the conventional FCN-32s head.
We do so because it is a better fit for our sparse hypercolumn pretraining, although the hypercolumn itself is built  using different layers (conv1 to conv5 and fc6 to fc7).
Thus the MLP predicting the embedding $\phi$ from hypercolumn features is discarded before finetuning for segmentation.
The training data consists of the PASCAL VOC 2012~\cite{pascal-voc-2012} train set augmented with annotations from~~\cite{hariharan11}.
Test results are reported as mean intersection-over-union (mIU) scores on the 
PASCAL VOC 2012 validation set (Column 6 of~\cref{accv18:t:pascalvoc}).
We are the state of the art on this benchmark among all self-supervised learning methods, even ones that use other supervisory signals than motion.

\subsection{Discussion}
\vspace{-0.1em}
We can take home several messages from these experiments.
First, in all cases our approach outperforms the baseline of predicting optical flow directly.
This supports our hypothesis that direct single-frame optical flow prediction is either too difficult due to its intrinsic ambiguity or a distraction from the goal of learning a good representation.
It also shows that our approach of predicting pairwise flow similarities successfully addresses this ambiguity and allows to learn good CNN representations from optical flow.

Second, our method is generally as strong as the state of the art for self-supervision using motion cues represented by the approach of~\cite{pathak17}.
In fact, our approach outperforms the latter by a large margin in PASCAL VOC07 classification.
This is notable as our approach is significantly simpler: by ingesting optical flow information directly, it does not require to pre-process the data via a video segmentation algorithm.

Finally, we also found out that all video/motion based methods for self-supervised learning are sometimes not as good as methods that use other cues. But our approach still sets the overall state of the art for semantic image segmentation suggesting that the learned representation may be more suitable for per-pixel applications.
This suggests that further progress in this area is possible and worth seeking.
At the same time,~\cite{doersch17} find that the \emph{combination} of different cues may in practice result in the best performance; in this sense, our approach, by significantly simplifying the use of motion cues, can make it much easier to design multi-task networks that can leverage motion together with other complementary methods.

\section{Conclusion}\label{accv18:s:conc}

We have presented a novel method for self-supervision using motion cues based on a cross-pixel optical-flow similarity loss function.
We trained an AlexNet model using this scheme on a large unannotated video data-set.
Visualizations of individual neurons in a deep layer and of the output embedding show that the representation captures structure in the image.

We established the effectiveness of the resulting representation by transfer learning for several recognition benchmarks.
Compared to the previous state of the art motion based method~\cite{pathak17}, our method
works just as well and in some cases noticeably better despite a significant algorithmic simplification.
We also outperform all other self-supervision strategies in semantic image segmentation (VOC12).
This is reasonable as we train on a per-pixel proxy task on undistorted RGB images.

Finally, we see our contribution as an instance of self-supervision using multiple modalities, RGB and optical flow, which poses our work as a special case of this broader area of research.
\section*{Acknowledgements}
The authors gratefully acknowledge the support of ERC IDIU, and the AWS Cloud Credits for Research program. The authors also thank Ankush Gupta and David Novotn\'{y} for helpful discussions and insights, and Christian Rupprecht, Fatma Guney and Ruth Fong for proof reading the paper.
\bibliographystyle{splncs}
\bibliography{refs}
\end{document}